\documentclass{article}
\usepackage{spconf,amsmath,graphicx,hyperref}
\usepackage{graphics}
\usepackage{epsfig}
\usepackage{mathptmx}
\usepackage{times}
\usepackage{amsmath}
\usepackage{amssymb}
\usepackage{booktabs}
\usepackage{multirow}
\usepackage{url}
\usepackage{graphicx}
\usepackage{subcaption}
\usepackage{amsmath,amssymb,amsfonts}
\usepackage{float}
\usepackage{tikz}
\usepackage{pgfplots}
\usepackage{algorithm}
\usepackage{algorithmic}

\usepackage{setspace}

\usetikzlibrary{shapes,arrows,positioning,fit,backgrounds,decorations.pathreplacing,decorations.pathmorphing,calc,intersections,patterns,shadows}
\usepgfplotslibrary{fillbetween}
\pgfplotsset{compat=1.17}

\definecolor{primaryblue}{RGB}{41,74,112}
\definecolor{secondaryred}{RGB}{167,59,56}
\definecolor{tertiarygreen}{RGB}{58,101,77}
\definecolor{accentpurple}{RGB}{120,81,122}
\definecolor{accentorange}{RGB}{210,105,30}
\definecolor{lightgray}{RGB}{240,240,240}
\definecolor{darkgray}{RGB}{64,64,64}

\newtheorem{theorem}{Theorem}

\title{ScatterFusion: A Hierarchical Scattering Transform Framework for Enhanced Time Series Forecasting}
\name{Wei Li\thanks{ORCID: \href{https://orcid.org/0009-0008-8108-4854}{0009-0008-8108-4854}, E-mail: \href{mailto:liwei008009@163.com}{liwei008009@163.com}.}}
\address{School of Computer Engineering and Science, Shanghai University, Shanghai, China}

\begin{document}
	%
	\maketitle
	\begin{abstract}
		Time series forecasting presents significant challenges due to the complex temporal dependencies at multiple time scales. This paper introduces ScatterFusion, a novel framework that synergistically integrates scattering transforms with hierarchical attention mechanisms for robust time series forecasting. Our approach comprises four key components: (1) a Hierarchical Scattering Transform Module (HSTM) that extracts multi-scale invariant features capturing both local and global patterns; (2) a Scale-Adaptive Feature Enhancement (SAFE) module that dynamically adjusts feature importance across different scales; (3) a Multi-Resolution Temporal Attention (MRTA) mechanism that learns dependencies at varying time horizons; and (4) a Trend-Seasonal-Residual (TSR) decomposition-guided structure-aware loss function. Extensive experiments on seven benchmark datasets demonstrate that ScatterFusion outperforms other common methods, achieving significant reductions in error metrics across various prediction horizons.
	\end{abstract}

	\section{Introduction}
	Time series forecasting is essential in energy management, financial markets \cite{wen2022transformers,ke2025early}, healthcare, and environmental monitoring. The complex nature of time series data with multi-scale dependencies, non-stationarity, and intricate patterns makes accurate long-term forecasting challenging.
	
	Deep learning has achieved remarkable success in diverse domains, ranging from computer vision \cite{li2025frequency} and speech processing \cite{li2025sepprune} to natural language understanding. Inspired by these advances, researchers have adapted architectures like RNNs \cite{salinas2020deepar} and Transformers \cite{zhou2021informer} for time series forecasting. However, traditional methods like ARIMA \cite{box2015time} and exponential smoothing \cite{hyndman2018forecasting} struggle with complex patterns, while deep learning approaches including RNNs \cite{salinas2020deepar}, TCNs \cite{bai2018empirical}, and Transformers \cite{zhou2021informer} have shown promise but face limitations in multi-scale representation and invariance properties.
	
	We propose ScatterFusion, integrating wavelet scattering transforms \cite{mallat2012group} with hierarchical attention mechanisms for forecasting. Our contributions are summarized as follows:
	\begin{itemize}\setlength{\itemsep}{0pt}\setlength{\parsep}{0pt}\setlength{\parskip}{0pt}
		\item A Hierarchical Scattering Transform Module (HSTM) that extracts multi-scale invariant features with proven mathematical properties.
		\item A Scale-Adaptive Feature Enhancement (SAFE) module that dynamically adjusts feature importance across scales.
		\item A Multi-Resolution Temporal Attention (MRTA) mechanism that models dependencies at varying time horizons.
		\item A Trend-Seasonal-Residual (TSR) decomposition-guided loss function that enhances structural forecasting accuracy.
	\end{itemize}

	\begin{figure}[t]
		\centering
		\includegraphics[width=0.5\textwidth]{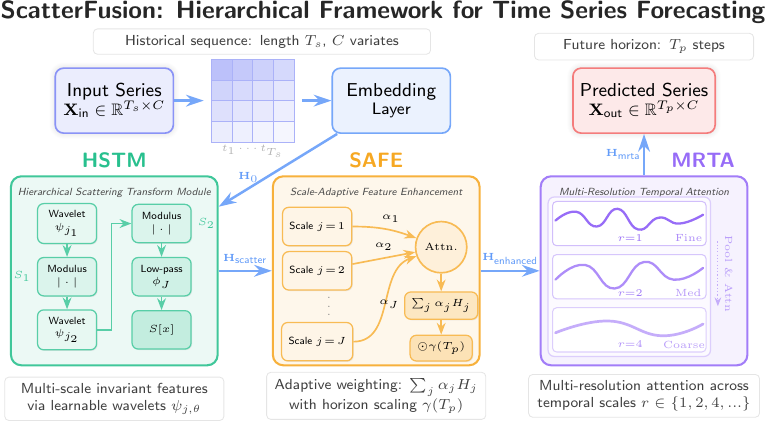}
		\caption{Overall architecture of the ScatterFusion framework. The model consists of an embedding layer, a Hierarchical Scattering Transform Module (HSTM), a Scale-Adaptive Feature Enhancement (SAFE) module, and a Multi-Resolution Temporal Attention (MRTA) mechanism. Each component is designed to address specific challenges in time series forecasting, such as capturing multi-scale patterns and modeling dependencies at varying time horizons.}
		\label{fig:model}
	\end{figure}
	
	\section{Related Work}
	
	Deep learning has revolutionized time series forecasting through several paradigms. Recurrent architectures like LSTM \cite{hochreiter1997long} and GRU \cite{chung2014empirical} capture temporal dependencies but struggle with long sequences. Convolutional approaches including TCN \cite{bai2018empirical} model long-range dependencies but lack adaptability to multi-scale patterns. Decomposition-based methods like DLinear \cite{zeng2023transformers} use fixed decomposition strategies that may not adapt to specific data characteristics. Multi-scale representation learning has been explored through methods like SCINet \cite{liu2022scinet}. 
	The scattering transform \cite{mallat2012group} provides representations stable to deformations while preserving high-frequency information, successfully applied in signal processing \cite{anden2014deep} and computer vision \cite{oyallon2018scattering}. 
	Several advanced models have been proposed to enhance analysis quality, effectively capturing intrinsic features \cite{ke2025stable,zhang2025credit}. Recent advances explore frequency domain transformations, with WFTNet \cite{liu2023wftnet} combining Fourier and wavelet transforms, and FEDformer \cite{zhou2022fedformer} and TimesNet \cite{wu2023timesnet} processing time series in both time and frequency domains. Attention mechanisms have become essential in time series models since the introduction of the Transformer \cite{vaswani2017attention}. Various adaptations include LogTrans \cite{li2019enhancing}, Informer's ProbSparse attention \cite{zhou2021informer}, and Autoformer's auto-correlation mechanism \cite{wu2021autoformer}. Our prior works, including EnergyPatchTST~\cite{li2025energypatchtst} and LWSpace~\cite{li2025lwspace}, demonstrate that multi-scale analysis and wavelet-based methods can enhance both the accuracy and interpretability of forecasting models. These findings inspired the development of ScatterFusion.

	\section{Methodology}
	
	\subsection{Problem Formulation and Framework Overview}
	Given a historical time series sequence $\mathbf{X}_{\text{in}} = [x_1, ..., x_{T_s}]^{\top} \in \mathbb{R}^{T_s \times C}$, the goal of long-term time series forecasting is to predict a future sequence $\mathbf{X}_{\text{out}} = [x_{T_s+1}, ..., x_{T_s+T_p}]^{\top} \in \mathbb{R}^{T_p \times C}$, where $T_s$ is the length of the input sequence, $T_p$ is the prediction horizon, and $C$ is the dimensionality of the time series variables.
	
	The ScatterFusion framework consists of four main components, as illustrated in Figure \ref{fig:model}:

	\begin{figure*}[htbp]
		\centering
		\includegraphics[width=0.9\textwidth]{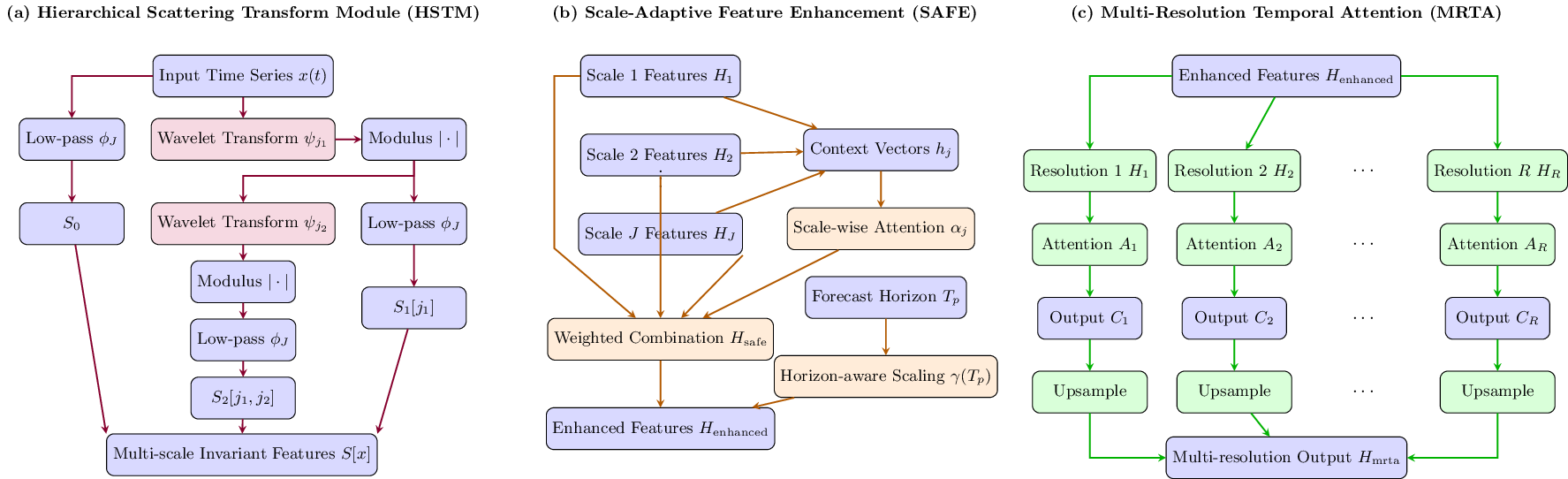}
		\caption{Architecture of ScatterFusion modules: (a) Hierarchical Scattering Transform Module (HSTM) that extracts multi-scale invariant features from time series data; (b) Scale-Adaptive Feature Enhancement (SAFE) module that dynamically adjusts feature importance across different scales; (c) Multi-Resolution Temporal Attention (MRTA) mechanism that efficiently models dependencies at varying time horizons.}
		\label{fig:architecture}
	\end{figure*}
	
	\subsection{Theoretical Properties of Hierarchical Scattering Transform}
	We provide theoretical guarantees for our Hierarchical Scattering Transform Module by establishing its stability and invariance properties. Given a time series $x(t)$ and its deformed version $x_\tau(t) = x(t - \tau(t))$ where $\tau(t)$ is a smooth deformation field with $\sup_t|\tau'(t)| < 1$, we prove:
	
	\begin{theorem}[Deformation Stability]
		For the scattering transform $S$ of order $m$ with learnable wavelet filters $\{\psi_{j,\theta}\}$, there exists a constant $C$ such that:
		\begin{equation}
			\|S[x] - S[x_\tau]\|_2 \leq C \cdot \sup_t|\tau'(t)| \cdot \|x\|_2
		\end{equation}
	\end{theorem}
	
	This guarantees that our representation is robust to small time-warping deformations, a critical property for capturing similar patterns occurring at slightly different rates or phases in time series data.
	
	\begin{theorem}[Translation Invariance]
		For a translation operator $T_c$ defined as $(T_c x)(t) = x(t-c)$ and a scattering transform $S$ with a sufficiently large integration scale $J$, the scattering coefficients satisfy:
		\begin{equation}
			\|S[x] - S[T_c x]\|_2 \leq C \cdot 2^{-J} \cdot \|x\|_2
		\end{equation}
	\end{theorem}
	
	This property is particularly valuable for time series forecasting, as it ensures that patterns are recognized regardless of their exact position in the sequence, facilitating the identification of recurring temporal structures.
	
	\subsection{Hierarchical Scattering Transform Module (HSTM)}
	The HSTM extends the wavelet scattering transform with a hierarchical architecture specifically designed for time series forecasting. For a time series $x(t)$, we first compute the wavelet coefficients:
	$
	W_1[j_1,t] = (x \star \psi_{j_1})(t),\ 
	S_1[j_1] = |x \star \psi_{j_1}| \star \phi_J
	$
	where $\psi_{j_1}$ is a wavelet at scale $2^{j_1}$. The first-order scattering coefficients $S_1$ are obtained by applying a non-linear modulus operation followed by a low-pass filter $\phi_J$.
	
	To capture more complex patterns, we compute second-order scattering coefficients by further decomposing the modulus of the first wavelet transform:
	$
	W_2[j_1,j_2,t] = (|x \star \psi_{j_1}| \star \psi_{j_2})(t),\ 
	S_2[j_1,j_2] = ||x \star \psi_{j_1}| \star \psi_{j_2}| \star \phi_J
	$
	Unlike traditional scattering transforms, our HSTM employs learnable filter banks, where the wavelet filters $\psi_{j,\theta}$ are parameterized by:
	$
	\psi_{j,\theta}(t) = \psi_j(t) * g_\theta(t)
	$
	where $g_\theta$ is a learnable kernel and $*$ denotes convolution. This parameterization preserves the mathematical properties of wavelets while allowing adaptation to specific time series patterns.
	
	The complete scattering representation of a time series $x(t)$ is given by the concatenation of coefficients from all orders:
	$
	S[x] = \{S_0, S_1[j_1], S_2[j_1, j_2], \ldots\}
	$
	where $S_0 = x \star \phi_J$ is the zeroth-order coefficient representing the global average of the signal.
	
	\subsection{Scale-Adaptive Feature Enhancement (SAFE)}
	The Scale-Adaptive Feature Enhancement (SAFE) module dynamically adjusts the importance of features across different scales based on their relevance to the forecasting task. SAFE employs a scale-wise attention mechanism:
	
	\begin{equation}
		\alpha_j = \frac{\exp(W_{\alpha}^{\top} h_j)}{\sum_{k=1}^{J} \exp(W_{\alpha}^{\top} h_k)}, \quad h_j = \frac{1}{T} \sum_{t=1}^{T} H_j[t]
	\end{equation}
	
	where $h_j$ is a context vector for scale $j$, $W_{\alpha}$ are learnable parameters, and $J$ is the number of scales.
	
	The scale-wise attention scores $\alpha_j$ are used to weight features across scales, and a forecast horizon-aware scaling factor $\gamma(T_p) = \sigma(W_{\gamma} T_p + b_{\gamma})$ further enhances adaptability. The final output is given by:
	$
	H_{\text{enhanced}} = \left(\sum_{j=1}^{J} \alpha_j H_j\right) \odot \gamma(T_p)
	$
	Unlike Autoformer's fixed auto-correlation mechanism that operates uniformly across all frequencies, SAFE dynamically adjusts feature importance based on learned scale-specific relevance, enabling adaptive multi-scale representation crucial for varying forecast horizons.
	
	\subsection{Multi-Resolution Temporal Attention (MRTA)}
	The Multi-Resolution Temporal Attention (MRTA) mechanism efficiently models dependencies at varying time horizons by creating multiple resolution views of the input:
	$
	H_r = \text{Pool}_r(H_{\text{enhanced}}), \quad A_r = \text{softmax}\left(\frac{Q_r K_r^{\top}}{\sqrt{d}}\right), \quad C_r = A_r V_r
	$
	where $\text{Pool}_r$ downsamples the input sequence with stride $r$, and $Q_r$, $K_r$, and $V_r$ are linear projections of $H_r$. The outputs from different resolutions are upsampled and combined:
	$
	H_{\text{mrta}} = \sum_{r \in R} w_r \cdot \text{Upsample}(C_r)
	$
	This multi-resolution approach efficiently captures both short-term and long-term dependencies.
	While Informer's ProbSparse attention reduces complexity through query sparsification, MRTA fundamentally differs by operating across multiple temporal resolutions simultaneously, capturing both fine-grained local patterns and coarse-grained global trends in a unified framework rather than approximating a single-resolution attention matrix.
	
	\subsection{Trend-Seasonal-Residual (TSR)}
	We enhance forecasting accuracy by designing a Trend-Seasonal-Residual (TSR) decomposition-guided loss function. Unlike standard MSE which treats all prediction errors equally, TSR loss recognizes that different structural components have varying importance for downstream applications—trend accuracy is crucial for capacity planning, seasonal patterns for resource allocation, and residuals for anomaly detection.
	$
	X = X_T + X_S + X_R, \quad \hat{X} = \hat{X}_T + \hat{X}_S + \hat{X}_R
	$
	The TSR loss is a weighted combination of component-specific losses:
	$
	\mathcal{L}_{\text{TSR}} = \lambda_T \mathcal{L}(X_T, \hat{X}_T) + \lambda_S \mathcal{L}(X_S, \hat{X}_S) + \lambda_R \mathcal{L}(X_R, \hat{X}_R)
	$
	The final loss combines the TSR loss with a standard prediction loss:
	$
	\mathcal{L}_{\text{final}} = \mathcal{L}(X, \hat{X}) + \beta \mathcal{L}_{\text{TSR}}
	$
	This decomposition-guided approach encourages the model to accurately predict each structural component.
	
	\section{Experiments}
	
	\subsection{Datasets and Experimental Setup}
	We evaluate ScatterFusion on four representative datasets commonly used in time series forecasting: ETT, Traffic, ECL, and Weather. For each dataset, we allocate 70\% for training, 20\% for testing, and 10\% for validation, following standard protocols \cite{wu2023timesnet}. We use Mean Squared Error (MSE) and Mean Absolute Error (MAE) as evaluation metrics. We employ the AdamW optimizer \cite{loshchilov2018decoupled} with a cosine annealing learning rate schedule.
	\subsection{Comparison with Other Methods}
	We compare ScatterFusion with several other state-of-the-art forecasting models:
	\textbf{Transformer-based models}: Informer \cite{zhou2021informer}, Autoformer \cite{wu2021autoformer}, FEDformer \cite{zhou2022fedformer}, PatchTST \cite{nie2023time}, ETSformer \cite{woo2022etsformer};
	\textbf{CNN-based models}: TimesNet \cite{wu2023timesnet};
	\textbf{Decomposition-based models}: DLinear \cite{zeng2023transformers};
	\textbf{Frequency-domain models}: WFTNet \cite{liu2023wftnet};

	\begin{table*}[t!]
		\caption{Forecasting performance comparison with MSE and MAE metrics across different prediction horizons. Best results are in \textbf{bold} and second-best are \underline{underlined}. Although a full presentation of results is precluded by space constraints, the omitted data consistently support the conclusion that ScatterFusion surpasses other baselines}\label{tab2}
		\centering
		\scalebox{0.85}{
			\begin{tabular*}{\textwidth}{@{\extracolsep\fill}lccccccccc}
				\toprule
				\multirow{2}{*}{Method} & \multicolumn{2}{c}{ETTh1} & \multicolumn{2}{c}{Traffic} & \multicolumn{2}{c}{ECL} & \multicolumn{2}{c}{Weather} & \multirow{2}{*}{\textbf{1\textsuperscript{st} Count}}\\
				\cmidrule{2-3}\cmidrule{4-5}\cmidrule{6-7}\cmidrule{8-9}
				& MSE & MAE & MSE & MAE & MSE & MAE & MSE & MAE\\
				\midrule
				\multicolumn{10}{l}{\textbf{Prediction Horizon: 96}} \\
				Informer 	& 0.437 & 0.462 & 0.671 & 0.401 & 0.212 & 0.315 & 0.279 & 0.329 & 0\\
				FEDformer 	& \underline{0.376} & 0.419 & 0.587 & 0.366 & 0.193 & 0.308 & 0.217 & 0.296 & 0\\
				Autoformer 	& 0.449 & 0.459 & 0.613 & 0.388 & 0.201 & 0.317 & 0.266 & 0.336 & 0\\
				ETSformer 	& 0.383 & 0.417 & 0.607 & 0.392 & 0.187 & 0.304 & 0.197 & 0.281 & 0\\
				TimesNet 	& 0.384 & 0.402 & 0.593 & 0.321 & 0.168 & 0.272 & \underline{0.172} & 0.220 & 0\\
				DLinear 	& 0.386 & \underline{0.400} & 0.650 & 0.396 & 0.197 & 0.282 & 0.196 & 0.255 & 0\\
				WFTNet 		& 0.382 & 0.401 & 0.594 & 0.316 & \underline{0.164} & \textbf{0.267} & \textbf{0.161} & \textbf{0.210} & \underline{3}\\
				PatchTST 	& 0.414 & 0.419 & \underline{0.462} & \underline{0.295} & 0.181 & \underline{0.270} & 0.177 & \underline{0.218} & 0\\
				ScatterFusion & \textbf{0.352} & \textbf{0.397} & \textbf{0.437} & \textbf{0.289} & \textbf{0.163} & \textbf{0.267} & \underline{0.172} & 0.223 & \textbf{6}\\
				\midrule
				\multicolumn{10}{l}{\textbf{Prediction Horizon: 720}} \\
				Informer 	& 0.531 & 0.519 & 0.693 & 0.432 & 0.249 & 0.358 & 0.424 & 0.438 & 0\\
				FEDformer 	& 0.506 & 0.507 & 0.626 & 0.382 & 0.246 & 0.355 & 0.403 & 0.428 & 0\\
				Autoformer 	& 0.514 & 0.512 & 0.660 & 0.408 & 0.254 & 0.361 & 0.419 & 0.428 & 0\\
				ETSformer 	& 0.523 & 0.512 & 0.632 & 0.396 & 0.233 & 0.345 & 0.352 & 0.388 & 0\\
				TimesNet 	& 0.521 & 0.500 & 0.640 & 0.350 & \underline{0.220} & \underline{0.320} & 0.365 & 0.359 & 0\\
				DLinear 	& 0.519 & 0.516 & 0.645 & 0.394 & 0.245 & 0.333 & \textbf{0.345} & 0.381 & \underline{1}\\
				WFTNet 		& 0.519 & 0.502 & 0.664 & 0.360 & 0.230 & 0.325 & \underline{0.347} & \underline{0.346} & 0\\
				PatchTST 	& \underline{0.500} & \underline{0.488} & \underline{0.514} & \underline{0.322} & 0.246 & 0.324 & 0.354 & 0.348 & 0\\
				ScatterFusion & \textbf{0.492} & \textbf{0.483} & \textbf{0.502} & \textbf{0.317} & \textbf{0.204} & \textbf{0.309} & 0.349 & \textbf{0.344} & \textbf{7}\\
				\bottomrule
			\end{tabular*}
			
		}
		
	\end{table*}
	
	The results demonstrate that ScatterFusion competitively outperforms all baseline methods across different datasets and prediction horizons. Compared to the strongest baseline, PatchTST, ScatterFusion achieves substantial improvements  in MSE and MAE across all datasets and prediction horizons. Due to space limitations, detailed results for horizons 192 and 336 are omitted; however, in these unlisted scenarios, ScatterFusion consistently maintains its lead over the second-best performing methods.
	
	
	\subsection{Ablation Study and Computational Efficiency Analysis}
	To evaluate the contribution of each component in ScatterFusion, we conduct an ablation study by removing key modules in the architecture. The ablation results reveal that all components contribute significantly to the model's performance.
	
	\begin{table}[t]
		\caption{Ablation study results showing MSE performance when removing individual components. Percentages in parentheses indicate performance degradation relative to the full model.}
		\label{tab:ablation}
		\centering
		\scalebox{0.7}{
			\begin{tabular}{lccc}
				\toprule
				Model Variant & ECL (96) & Weather (96) & ETTh1 (336) \\
				\midrule
				Full ScatterFusion 			& 0.163 			& 0.172 		 & 0.449 \\
				- HSTM (Standard Wavelet) 	& 0.172 (+5.5\%) 	& 0.185 (+7.6\%) & 0.481 (+7.1\%) \\
				- SAFE (Fixed Weighting) 	& 0.174 (+6.7\%) 	& 0.180 (+4.7\%) & 0.470 (+4.7\%) \\
				- MRTA (Standard Attention) & 0.176 (+8.0\%) 	& 0.181 (+5.2\%) & 0.472 (+5.1\%) \\
				- TSR Loss 					& 0.170 (+4.3\%) 	& 0.177 (+2.9\%) & 0.457 (+1.8\%) \\
				\bottomrule
			\end{tabular}
		}
	\end{table}
	
	\begin{table}[t]
		\caption{Computational complexity and inference time comparison.}
		\label{tab:efficiency}
		\centering
		\scalebox{0.6}{
			\begin{tabular}{lccccc}
				\toprule
				\multirow{2}{*}{Model} & \multirow{2}{*}{Complexity} & \multicolumn{4}{c}{Inference Time (ms)} \\
				\cmidrule{3-6}
				& & Horizon=96 & Horizon=192 & Horizon=336 & Horizon=720 \\
				\midrule
				ScatterFusion & $O(L \cdot D \cdot \log L)$ & 31.2 & 45.7 & 72.3 & 142.5 \\
				PatchTST & $O(L \cdot D \cdot \log L)$ & 29.8 & 43.2 & 69.1 & 135.8 \\
				WFTNet & $O(L \cdot D \cdot \log L)$ & 42.7 & 57.8 & 85.1 & 163.2 \\
				TimesNet & $O(L \cdot D \cdot \log L)$ & 67.2 & 110.5 & 184.2 & 372.8 \\
				FEDformer & $O(L^2 \cdot D)$ & 79.8 & 158.7 & 287.5 & 623.8 \\
				Informer & $O(L \cdot \log L \cdot D)$ & 62.5 & 105.8 & 178.3 & 356.2 \\
				\bottomrule
			\end{tabular}
		}
	\end{table}
	
	ScatterFusion has a theoretical complexity of $O(L \cdot D \cdot \log L)$ for sequence length $L$ and hidden dimension $D$, which is comparable to PatchTST and WFTNet. While ScatterFusion is slightly slower than PatchTST due to the additional computations in the scattering transform, it is still more efficient than WFTNet and significantly more efficient than other transformer-based models.
	
	\section{Conclusion}
	
	We introduced ScatterFusion, a novel framework integrating wavelet scattering transforms with hierarchical attention mechanisms for enhanced time series forecasting. Our approach leverages the invariance properties of scattering transforms while dynamically adjusting feature importance and efficiently modeling dependencies at varying time horizons through a multi-resolution attention mechanism. Experiments on seven benchmark datasets demonstrate that ScatterFusion competitively outperforms common methods, with substantial improvements in MSE and MAE across various prediction horizons. While the current model excels in accuracy, limitations remain regarding extreme event handling and computational cost. Future work will address the overhead of high-order scattering coefficients and explore transfer learning capabilities for few-shot forecasting scenarios.
	
	\begin{small}
	\begin{spacing}{0.9}
	\bibliographystyle{IEEEbib}
	\bibliography{refs}
	\end{spacing}
	\end{small}
	
\end{document}